\documentclass[accepted]{uai2026} 

\usepackage{tcolorbox}
\usepackage{threeparttable}
\usepackage{paralist}

\usepackage[american]{babel}

\usepackage{natbib} 
\usepackage{mathtools} 
\usepackage{siunitx} 
\usepackage{booktabs} 
\usepackage{tikz} 
\usepackage{caption}
\usepackage{multirow}
\usepackage{subcaption}
\usepackage{amssymb}
\usepackage{amsthm}

\newcommand{\ccumort}{\textbf{CCU-Mort}}

\newtheorem{proposition}{Proposition}[section]
\newtheorem{theorem}{Theorem}[section]
\newtheorem{definition}{Definition}[section]
\newtheorem{remark}{Remark}[section]
\newcommand{\titleName}{Aligning Probabilistic Beliefs under Informative Missingness:\\LLM Steerability in Clinical Reasoning}

\title{\titleName}

\author[1]{\href{mailto:<yk3043@cumc.columbia.edu>?Subject=Mind the data gap}{Yuta Kobayashi$^*$}{}}
\author[1]{Vincent Jeanselme\footnote{Equal contribution}}
\author[1]{Shalmali Joshi}
\affil[1]{%
    Department of Biomedical Informatics
    Columbia University
    New York City
}
\begin{document}

\maketitle

\begin{abstract}
Large Language Models (LLMs) are increasingly deployed for clinical reasoning tasks, which inherently require eliciting calibrated probabilistic beliefs based on available evidence. However, real-world clinical data are frequently incomplete, with missingness patterns often informative of patient prognosis; for example, ordering a rare laboratory test reflects a clinician's latent suspicion. In this work, we investigate whether LLMs can be steered to leverage this informative missingness for prognostic inference. To evaluate how well LLMs align their verbalized probabilistic beliefs with an underlying target distribution, we analyze three common prompt-based interventions: explicit serialization, instruction steering, and in-context learning. We introduce a bias-variance decomposition of the log-loss to clarify the mechanisms driving gains in predictive performance. Using a real-world intensive care testbed, we find that while explicit structural steering and in-context learning can improve probabilistic alignment, the models do not natively leverage informative missingness without careful interventions.
\end{abstract}

\begin{figure*}[!ht]
    \centering
    \includegraphics[width=.8\linewidth]{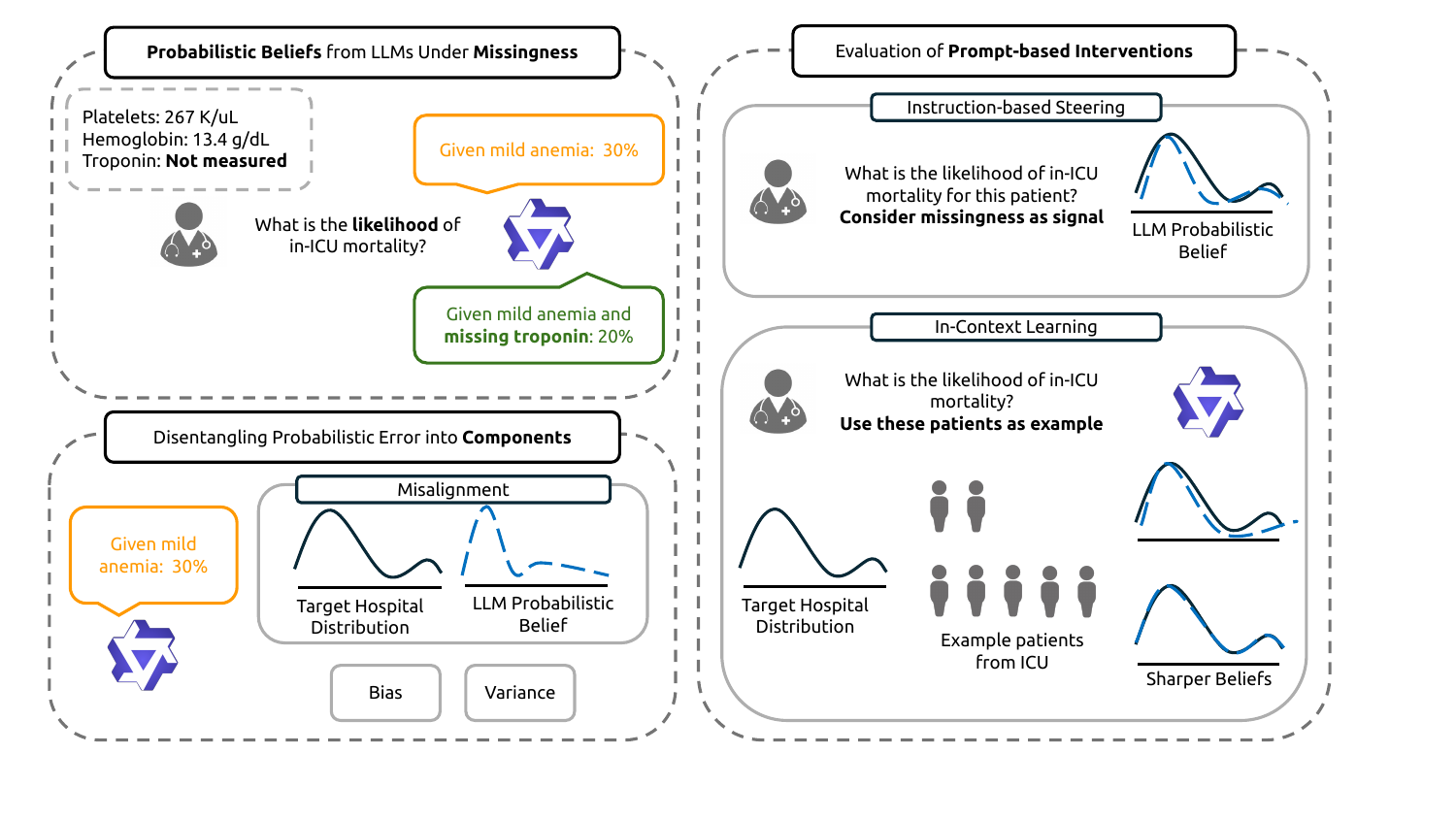}
    \caption{Our work shows that explicit instructions to consider missing information affect predictive performance and verbalized probabilistic estimates.} 
    \label{fig:summary}
\end{figure*}

\section{Introduction}
From question answering to diagnosis, Large Language Models (LLMs) have the potential to transform patient care and inform clinical decision-making. In particular, these models are increasingly evaluated as clinical reasoning tools, relying on encoded medical knowledge priors in pretraining data to inform their outputs using text-serialized medical context~\citep{makarov2025large, hegselmann2023tabllm, requeima2024llm, su2025multimodal}. With increasing interest in using these tools to inform diagnosis and prognosis~\citep{mcduff2025towards, shahsavar2023user, sellergren2025medgemma}, initial results show promising LLM performance for patient prognosis~\cite{helmy2025leveraging, cui2025llms, zhu2024prompting, chen2025narrative}. 

However, clinical prognosis requires LLMs to reason from partially observed data, introducing two challenges. 
First, these partial observations inherently reflect complex interactions between patients and the healthcare system \citep{sisk2021informative, tan2023informative, jeanselme2024clinical}. Crucially, the choice of collected measurements provides insights into a patient's condition, available resources, and provider decision-making. For example, in the Intensive Care Unit (ICU), the mere presence (or absence) of a measurement often reflects a clinician's latent suspicion (or lack of suspicion) of risk just as much as the value of the measurement itself. Essentially, missingness in healthcare tends to be informative. Machine learning practitioners routinely leverage these patterns as proxies for patient acuity, and including indicators for unmeasured tests has been shown to significantly enhance the predictive performance of neural networks \citep{lipton2016modeling, che2018recurrent}. Importantly, discarding these informative patterns of missingness does not resolve the issue and may lead to biased risk estimates~\citep{agniel2018biases}.

Second, LLMs' output must reflect the true predictive uncertainty aligned with the underlying data. That is, for an LLM to effectively support downstream clinical decision-making, such as patient triage or management, the predicted risk must be calibrated to the target distribution~\citep{nizri2025does}. Without this alignment, the model cannot safely inform decision-making~\citep{amodei2016concrete} or further reason about potential interventions, such as which treatments to prioritize or what additional data to acquire to resolve uncertainty.

LLMs naturally accommodate incomplete information through natural language, as a user can include only observed measurements in the input. However, when the underlying missingness patterns are informative of the patient's state, they inform clinicians' reasoning and beliefs about the patient. LLMs' broader ability to reason about the underlying patterns of missingness in the context of clinical tasks has not been systematically studied. More generally, the understanding of informative missingness and its impact on uncertainty quantification is limited to the traditional task-specific supervised models trained on labeled data with a given outcome. It remains unclear whether LLMs encode prior knowledge of the correlations between potentially informative missingness and outcomes, and, if so, how this knowledge may inform verbalized probabilistic beliefs when prompted with partially observed information. Our work aims to fill this gap by measuring the impact of informative missingness on LLMs' uncertainty quantification. Further, we develop prompt-based steering strategies to align LLMs' probabilistic beliefs with the underlying risk distribution.

We formalize the problem of `reasoning' about informative missingness, in the context of common end-user prompting interventions widely utilized to improve LLM performance: (i) naive serialization, (ii) instruction steering, and (iii) in-context learning (ICL). As illustrated in Figure~\ref{fig:summary}, we analyze the effectiveness of these interventions in eliciting calibrated probabilistic beliefs that actively leverage informative missingness. Specifically, we introduce a bias-variance decomposition of the LLM's predictive error. This theoretical framework clarifies the underlying mechanisms that may drive performance improvements, such as increasing model size, domain-specific fine-tuning, and increasing context sample sizes. Our analysis informs steering interventions to align verbalized beliefs based on informative missingness and evaluate their impact on Electronic Health Records (EHRs)-based mortality prediction in the MIMIC-IV dataset \citep{johnson2020mimic}.

We find that in zero-shot settings, LLMs are unable to natively leverage explicit missingness indicators or infer their correlations with outcomes from prior encoded knowledge. However, by providing a structural steering instruction, the models can effectively adjust their probabilistic beliefs. Additionally, while In-Context Learning (ICL) enables LLMs to align their risk estimates with target distributions, we identify critical failure modes in which in-context samples induce predictive overconfidence.

Together, these findings lead to the following conclusions:
\begin{asparaitem}
\item Instructions can be utilized to inject structural prior knowledge about missingness into an LLM's probabilistic reasoning. While out-of-the-box models fail to natively capture the mechanisms of informative missingness, explicitly instructing these constraints allows practitioners to steer the model's probabilistic beliefs.
\item We demonstrate that while $k$-shot ICL is effective at aligning a model's risk estimates with a target distribution, it lacks inherent regularization in complex predictive tasks, such as clinical risk assessment.
\end{asparaitem}

\section{Related work}

\paragraph{LLMs and predictive uncertainty.}
To support safe decision-making~\citep{amodei2016concrete}, LLMs' probabilistic belief, e.g. verbalized risk estimate, must accurately reflect the underlying outcome distribution, requiring both reliable uncertainty estimation and its alignment with the underlying data-generating process.

Uncertainty estimates have been obtained using various approaches. For example, token-level perplexity/uncertainty~\citep{malinin2020uncertainty}, verbalized uncertainty generated by LLMs~\citep{krause2023confidently, mielke2022reducing, lin2022teaching, tian2023just}, and ensembling model predictions~\citep{hou2023decomposing}. Critically, decoding mechanisms introduce algorithmic stochasticity that may obscure the relationship between an LLM's posterior predictive distribution~\citep{hashimoto2025decoding, shi2024thorough} and the underlying uncertainty it encodes. To mitigate this issue, self-consistency quantification~\citep{wang2022self} or sampling from the posterior predictive distribution is often used~\citep {hagele2026hotmess, taubenfeld2025confidence}. 

Ensemble~\citep{zhang2024luq}, in-context learning~\citep{ling2024uncertainty, zhang2024study}, fine-tuning~\citep{mielke2022reducing, kapoor2024calibration}, and reinforcement learning~\citep{xu2024sayself} approaches have been proposed to close the gap between predicted and observed distributions, i.e., to align beliefs with the underlying data distribution. We focus on aligning verbalized beliefs under informative missingness motivated by end-user needs. Specifically, using sampled verbalized uncertainty quantification~\citep{hagele2026hotmess}, we aim to align LLMs' probabilistic beliefs under informative missingness through prompt-based interventions. Closest to our work, \citet{hagele2026hotmess} introduces a bias-variance decomposition of predictive error to motivate an ensemble correction for the variance in models' outputs associated with complex tasks. We use a similar decomposition to motivate three prompt-based interventions: serialization, prompt steering, and ICL.

\paragraph{Missingness and LLMs.}
Because natural language rarely presents missing tokens, the problem of missingness has often been ignored in LLM pretraining. At inference time, the capacity of LLMs to predict the next token has been leveraged to impute missing values, such as infilling blank text~\citep{donahue2020enabling}, time series completion~\citep{gruver2023large}, or missing measurements or values~\citep{ding2024data, he2024llm, nazir2023chatgpt}. Such approaches aim to leverage domain-specific priors to improve imputation. Our work takes a different stance: rather than eliminating missingness, we treat it as a potentially informative signal and study how its representation shapes LLMs' predictive posteriors.
Two works are closest to this perspective. \citet{fu2025absencebench} demonstrates that LLMs fail to identify missing text when prompted to compare two short vignettes. From their experiments, LLMs are not natively able to identify missing information. By adding placeholders for missing data, LLMs show improved performance. This observation motivates our analysis of serialization. \citet{wang2024LLMs} proposes an ask-before-answer approach using a chain-of-thought to identify which missing information should be acquired before answering a question. Together, these works establish that LLMs can reason about potentially missing information through explicit interventions. Our work explores whether such reasoning capacity may be leveraged to improve clinical prognosis.

\paragraph{LLMs for EHR prognoses.} LLMs' zero- and few-shot capabilities~\citep{brown2020language} offer a compelling path for clinical prediction, where labeled data are expensive to collect. By interfacing with structured Electronic Health Records (EHRs) via text serialization \citep{hegselmann2025large}, LLMs have been applied to tasks ranging from structured data retrieval \citep{agrawal2022large, sellergren2025medgemma} to complex diagnostics conditioned on historical patient cases \citep{xiao2025retrieval, zhou2025large}, including medical outcome predictions such as readmission risk and at-risk patient identification \citep{liu2023large, labrak2024zero, helmy2025leveraging, cui2025llms, chen2025narrative}. Despite these advances, the existing literature rarely evaluates LLMs' capacity to produce calibrated probabilistic beliefs --- an important prerequisite for informing further reasoning and downstream clinical decision-making, such as triage and treatment prioritization. Critically, EHRs are often a partial window into patients' health, highlighting a gap in the literature: \textit{how does missingness influence LLM predictive uncertainty for clinical prognosis?}

\section{Steering LLMs under informative missingness}
\label{sec:method}

Consider $X\in\mathcal{X} = \mathbb{R}^d$ to be features, $M\in \mathcal{M}=\{0,1\}^d$, the missingness indicators, and $Y\in\{0,1\}$, the binary outcomes. Let $\mathcal{Y}$ denote the probability simplex. We use $X_{\text{obs}} = X \odot M$ to represent the observed features. We define informative missingness as $M \not \perp\!\!\!\!\perp Y \mid X_{\text{obs}}$ (see proof of when this phenomenon occurs in App.~\ref{app:info_mnar}). In this context, our probabilistic inference task is to model the conditional distribution of outcomes $Y$ given observed measurements $X_{\text{obs}}$ and their missingness patterns $M$.

We treat a pretrained LLM as a static functional hypothesis class that maps inputs to probability distributions over the output space. Additionally, the choice of prompt determines which function is selected from this space, as different prompts instantiate different predictive behaviors. Concretely, we define $\mathcal{Q}$ as the discrete set of functions $q: \mathcal{X} \times \mathcal{M} \times \mathcal{P} \rightarrow \mathcal{Y}$ reachable by the frozen LLM under any prompt configuration $\psi \in \mathcal{P}$, producing, under a given decoding strategy, a verbalized probabilistic belief in the output space.

We measure the capacity of an LLM to reason, defined as its capacity to update its probabilistic beliefs given input, through the \emph{expected} verbalized probability estimate of an LLM. As a decoding policy $\pi$ (with temperature $\tau > 0$) is a stochastic process, we define the expected verbalized probability estimate for a prompting parameter $\psi$ by marginalizing out the decoding stochasticity of the LLM: 
\begin{align*}
    q_\psi &= \int q(Y \mid X_{\text{obs}}, M, \psi, \varepsilon) p(\varepsilon) d\varepsilon
\end{align*}
In which $\varepsilon \sim p(\varepsilon)$ captures the decoding stochasticity. 

\begin{remark}
    We focus our analysis on the expected verbalized probability distribution as it represents the LLM's verbalized belief after integrating out the stochasticity of the decoding process. This provides a robust proxy for the model's underlying reasoning capabilities. It balances the reality of how these models are queried (via a single sample) with the theoretical need to evaluate the stability of the model's probabilistic output independent of a specific sample of $\varepsilon$.
\end{remark}

Let $p^*(Y \mid X_{\text{obs}}, M)$ denote the true probabilistic mechanism under the data-generating distribution. Because the LLM's weights and pretraining data are fixed, $\mathcal{Q}$ is static. Our aim is to obtain $q^*$, the best approximation of $p^*$ in $\mathcal{Q}$.

\begin{definition}[Projection of $p^*$] We define the parameter $\psi^*\in\mathcal P$ as the minimizer of the Kullback-Leibler divergence from the true distribution $p^*$ to the model family: 
\begin{align*}
\mathop{\mathrm{inf}}_{q \in \mathcal{Q}} \mathbb{E}_{X, M}\big[\mathrm{KL}(p^* \,\|\, q(Y \mid X_{\text{obs}} , M, \psi))\big].
\end{align*}
We denote the corresponding optimal predictor as $q^*$. 
\label{def:projection}
\end{definition}
Note that we do not assume that the true target distribution is representable in $\mathcal{Q}$. If it does, then $q^* = p^*$. Otherwise, the $q^*$ results in the irreducible lower bound on the estimation error (the Prior Misalignment) given the model and its pretraining constraints. Informally, $q^*$ is the orthogonal projection of the true distribution $p^*$ onto the space of distributions representable by $\mathcal{Q}$. 

To formalize the ability of LLMs to elicit probabilistic beliefs from informative missingness, we consider common end-user prompting interventions that leverage the information users can provide to reduce the LLM's approximation error: serialization, instruction-based steering, and in-context learning. Therefore, we define a specific prompting configuration as the tuple $\psi = (s, I, C_k)$, where each component is defined as follows.  

\paragraph{Serialization.} Let the serialization function be $s : \mathcal{X} \times \mathcal{M} \rightarrow \mathcal{S}$ as the mapping from the raw feature-missingness pair to the token sequence provided to the model. We contrast two dominant strategies: 
\begin{asparaitem}
    \item Implicit Serialization ($s_{\text{imp}}$): Only observed features are tokenized; missing values are dropped.
    \item Explicit Serialization ($s_{\text{exp}}$): Missing values are explicitly tokenized as distinct placeholders ("Not measured").
\end{asparaitem}

\paragraph{Instructions.} Instructions \citep{ouyang2022training, yuksekgonul2024textgrad, akinwande2023understanding} have often been used as a method for steering LLMs. We formalize the finite set of possible instructions $\mathcal I$ where an instruction $I \in \mathcal I$ constrains the reachable space of predictive functions within the static hypothesis space $\mathcal{Q}$. Concretely, we define a restricted prompt family $\mathcal{P}_I = \{(\phi, I, C) \in \mathcal{P} \mid I \in \mathcal I\}$ which fixes the structural instruction to $I$ while allowing context samples to vary, inducing a reachable space $\{ q_\psi : \psi \in \mathcal{P}_I \}$.

\paragraph{In-context learning (ICL).} Let $C_k = \{(X_{\text{obs}, i}, M_i, Y_i)\}_{i=1}^k$ denote a set of $k$ independent context examples provided in the prompt, sampled from $p^*$. Note that all context examples are serialized according to $s$. We hypothesize that these context samples provide the LLM with empirical evidence about the target distribution. Informally, we can view this process as analogous to a posterior update: as the context size $k$ increases, the model "sharpens" its internal belief state, reducing the entropy of its predictive distribution over the hypothesis space.

\begin{remark}
    We refrain from formalizing LLM approximations as exact Bayesian inference. Theoretical work models ICL as implicit Bayesian updates over a predefined latent task prior, assuming the model was trained and evaluated on tasks drawn from that distribution \citep{xie2021explanation,ye2024exchangeable,wakayama2025context}.  Recent studies suggest that out-of-the-box LLMs trained on massive, heterogeneous corpora, deviate significantly from optimal Bayesian behavior \citep{arora2024bayesian, falck2024context}. Therefore, we adopt a more general, but functional perspective: rather than a posterior update, the context $C_k$ serves as a step toward the optimal projection $q^*$.
\end{remark}

We analyze the alignment between the LLM probabilistic belief and the underlying process, i.e., the expected KL divergence between $p^*$ and the ICL predictor $q_{\psi}$ induced by a given choice of prompting strategy $\psi = (s, I, C_k)$, while keeping $C_k$ stochastic, and show that the expected risk decomposes into two distinct sources of error. We note that in the context of cross-entropy loss, this decomposition maps directly to the Bias-Variance Decomposition \citep{heskes1998bias, domingos2000unified, hagele2026hotmess}. 

\begin{theorem}[Error decomposition]
\label{theorem:decomposition}
Let $\mathcal{L}(q_{\psi}) = \mathbb{E}_{X,M,C_k}\big[ \mathrm{KL}(p^* \,\|\, q_{\psi}) \big]$ be the expected KL divergence of the marginalized predictor induced by $\psi = (s, I)$, integrated over the feature space $\mathcal{X} \times \mathcal{M}$ and the sampling distribution of the context $C_k$. We have the following decomposition: 
\begin{align*} \mathcal{L}(q_{\psi}) = & \underbrace{\mathbb{E}_{X,M}\Big[ \mathrm{KL}\big(p^*  \mid\mid q^*\big) \Big]}_{\text{Bias: Prior Misalignment}} \\
& + \underbrace{\mathbb{E}_{X,M}\Big[\mathbb{E}_{C_k}\big[\mathrm{KL}\big(q^* \mid\mid q_{\psi}\big)\big] \Big]}_{\text{Variance: Estimation Error}} \end{align*}
\end{theorem}

The decomposition clarifies distinct mechanisms in which the LLM's approximation error may be reduced. The bias term is irreducible given a fixed LLM, but increasing model capacity or finetuning may lead to improvements. Providing information within the prompting strategy $\psi = (s, I, C_k)$ using explicit missingness indicators, informative instructions along with context samples can better "align" the LLM's approximation with the optimal $q^*$. We now introduce measures of the gain achieved by these strategies.

\begin{definition}[Representation Gain] Let $q_{s_{\text{imp}}}$ be the predictor under implicit serialization. The Representation Gain of explicit serialization $s_{\text{exp}}$ is the reduction in expected divergence relative to the implicit baseline:$$\mathcal{R}(s_{\text{exp}}) := \mathbb{E}_{X,M}\Big[ \mathrm{KL}(p^* \,\|\, q_{s_{\text{imp}}}) - \mathrm{KL}(p^* \,\|\, q_{s_{\text{exp}}}) \Big]$$\end{definition}

A positive representation gain $\mathcal{R} > 0$ implies that serializing missingness as input tokens enables the LLM to attend to this process, effectively steering the predictor in a subspace of $\mathcal{Q}$ that includes functions dependent on $M$.

\begin{definition}[Steering Gain] Let $q_I$ and $q_\emptyset$ be the predictor realized by the model with and without structural instructions, respectively. The Steering Gain of an instruction $I$ is the reduction in the expected KL divergence from the true distribution $p^*$ relative to this baseline:$$\mathcal{S}(I) := \mathbb{E}_{X,M}\Big[ \mathrm{KL}(p^* \,\|\, q_\emptyset) - \mathrm{KL}(p^* \,\|\, q_I) \Big]$$\end{definition}

We hypothesize that the addition of structural instructions prunes the space of reachable functions, leading to reduced variance in the realized predictor. Without these instructions, the model requires a large context size $k$ to converge toward the optimal $q^*$. Therefore, the Steering Gain $\mathcal{S}(I)$ may also be seen as a reduction in sample complexity when combined with ICL: a steered model requires fewer examples ($k_{\text{steered}} \ll k_{\text{baseline}}$) to achieve the same approximation error (Cross-Entropy loss) relative to $q^*$. We provide a brief discussion for when such sample complexity improvements may be achieved in App.~\ref{app:steering_theory}. Importantly, note that instructions can lead to negative $\mathcal{S}(I)$, corresponding to choices that steer predictive beliefs further from $q^*$ or even constrain the space of reachable functions to exclude $q^*$.

\section{Experimental Design}
We construct an experiment to evaluate\footnote{Code available at \url{https://github.com/reAIM-Lab/EHR-missingness/}} the efficacy of prompt-based interventions in eliciting calibrated probabilistic beliefs from LLMs. Our experimental design directly mirrors our theoretical error decomposition. Specifically, we isolate the irreducible prior misalignment by evaluating how varying representational capacity affects baseline performance. In addition, we analyze the reducible estimation error and quantify how structural instruction steering and increased in-context sample sizes systematically mitigate it.

\paragraph{Real-world data.} For our empirical evaluation, we analyze in-hospital mortality (\ccumort) prediction from laboratory tests for a cohort of $N=500$ patients following an emergency admission to the Coronary Care Unit (CCU) from the MIMIC-IV database \citep{johnson2020mimic}. The observation window is restricted to the first 48 hours after admission, and a feature is considered missing if no measurement is recorded during this period. This task choice is motivated by the potential informativeness of missingness in CCU settings, where diagnostic orders are indicators of a clinician's latent suspicion of deterioration. For instance, arterial blood gas (ABG) measurements are invasive and typically reserved for acute respiratory distress, while serial lactate or white blood cell (WBC) tests strongly signal suspected sepsis or hypoperfusion. Preliminary experiments using a logistic regression baseline confirm this hypothesis, as including the missingness mask $M$ yields improvement in log-loss as shown in App.~\ref{app:lr_baseline}. Our setting aims to measure how these missingness patterns may influence LLMs' predictions. Note that including additional features or modalities can improve predictive performance, but render the missingness signal less informative. In the absence of those, this experiment evaluates LLMs' ability to appropriately calibrate beliefs to improve predictive performance. 

\paragraph{Models.} To investigate the effect of representational capacity via increasing model size, we evaluate the Qwen-3 family of models \citep{qwen3technicalreport} across four parameter sizes (4B, 8B, 14B, and 32B). We also measure prior alignment by comparing zero-shot evaluations of Gemma and MedGemma 27B models \citep{team2025gemma, sellergren2025medgemma}. 

\begin{figure*}[!ht]
    \centering
    \setlength{\tabcolsep}{1pt} 
    \begin{tabular}{ccc}
        \includegraphics[height=140px]{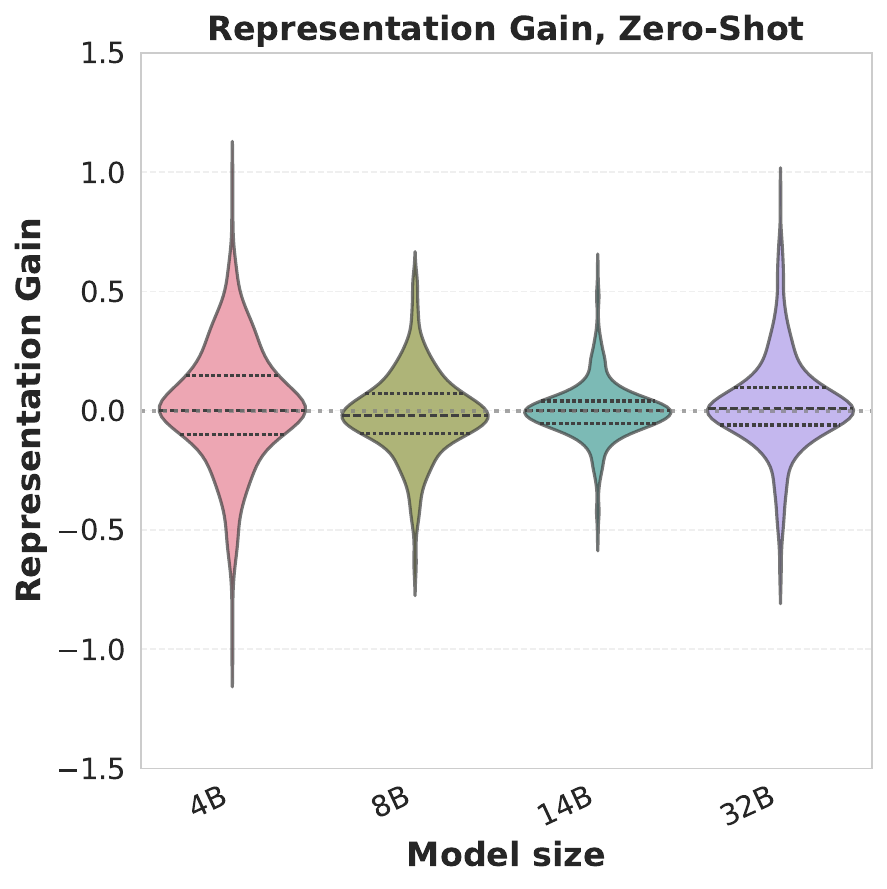} & 
        \includegraphics[height=140px]{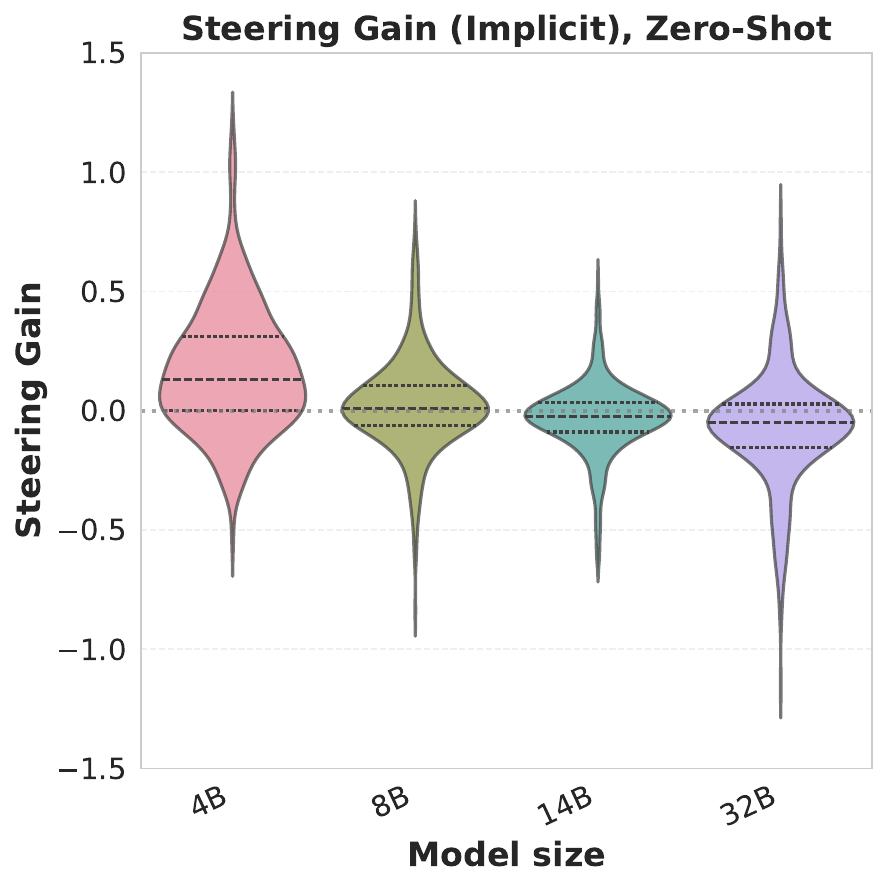} &
        \includegraphics[height=140px]{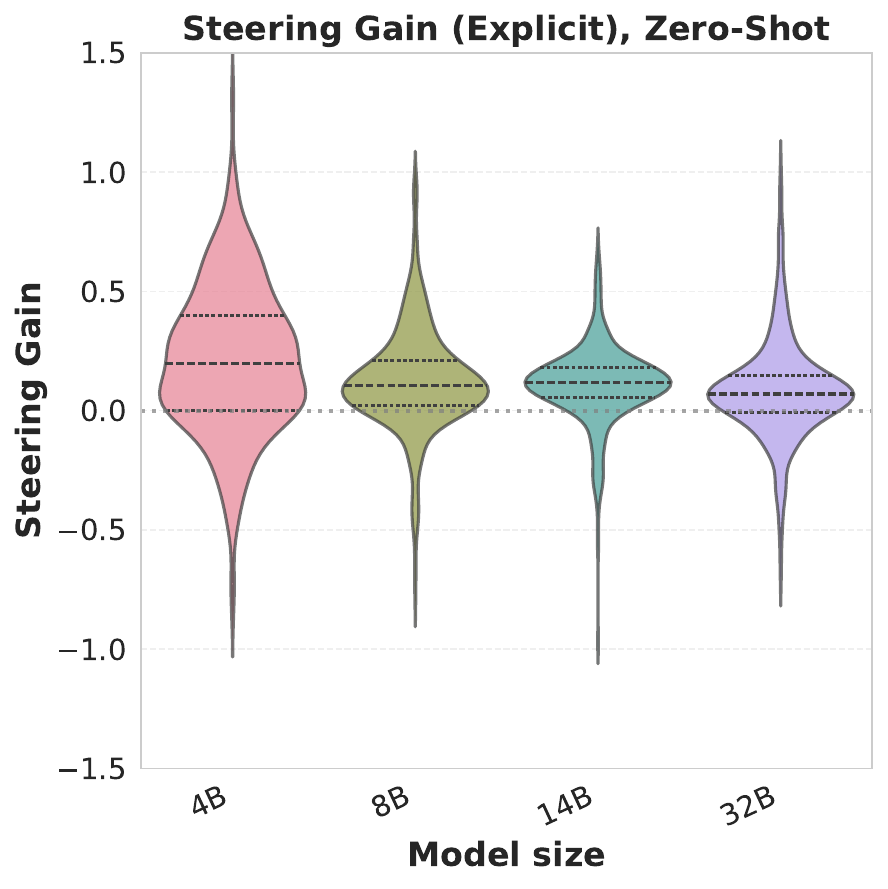} \\
        \small (a) Representation gain & \small (b) Implicit steering gain & \small (c) Explicit steering gain  \\
    \end{tabular}
    \caption{Impact of missingness serialization and instruction steering on individual log-loss difference as a function of model size for zero-shot inference. \textit{Positive values correspond to improved log-loss under the intervention compared to the base prompt.} }
    \label{fig:mimic:serialization:logloss}
\end{figure*}

\paragraph{Base prompt.} Our prompt consists of a serialization of continuous measurements and task-specific instructions. For serialization, we enumerate all selected laboratory tests in a textual format, as proposed by~\cite{hegselmann2025large} and adopted in various subsequent works \citep{lee2025clinical} (see App.~\ref{app:prompt} for a detailed example). For each task, we query the model to quantify the risk of the condition and end the query with a formatting instruction to output a verbalized estimate of risk as a probability between 0.0 and 1.0 ~\citep{kadavath2022language, lin2022teaching, kapoor2024large}.

\paragraph{Evaluation.} For all settings, we obtain $5$ samples per inference subject with temperature of $0.7$. For ICL, we also sample $5$ different context sets, yielding $25$ verbalized predictions. To evaluate the quality of the probabilistic beliefs, we compute expected calibration error (ECE) to measure how well predicted risks align with the true test distribution, as well as log-loss. ECE assesses whether the model's estimated probabilities empirically match observed frequencies at the population level, and log-loss quantifies how well the model's probability distribution is concentrated around the true outcome for each sample, while allowing us to assess quantities defined in Section~\ref{sec:method} (see how representation and steering gains are estimated in App.~\ref{app:test_statistic}). The following focuses on log-loss; ECE is deferred to App.~\ref{app:tab_res}.

\section{Results}
Our empirical evaluation proceeds in two stages. First, in Sections \ref{sec:zero-shot-serialization} and \ref{sec:zero-shot-inst}, we evaluate representation gain and steering gain in a strictly zero-shot setting across model sizes on \ccumort. Second, we analyze the impact of in-context learning (ICL) in Section \ref{sec:few-shot-serialization}: we examine ICL as a mechanism for learning general patterns in the target distribution, then we explore its interaction with instruction-based steering to determine if ICL improves the LLM's ability to leverage missingness as a predictive signal. Throughout, we use a Logistic Regression model with the full cohort ($n=2699$) as the baseline (Table~\ref{tab:lr_full} in Appendix describes performance as the number of samples increases, with and without indicators of missingness).

\subsection{Serialization: Zero-shot LLMs Fail to Leverage Explicit Missingness}
\label{sec:zero-shot-serialization}

\paragraph{Intervention.} This first experiment compares LLMs using the base prompt, in which missingness is dropped (denoted as \textbf{Dropped} strategy), with a prompt in which all features that are not measured over the 48 hours post-ICU admission are indicated as "Not measured", which we denote as the \textbf{Indicator} strategy. 

\begin{figure*}[!ht]
    \centering
    \includegraphics[height=140px]{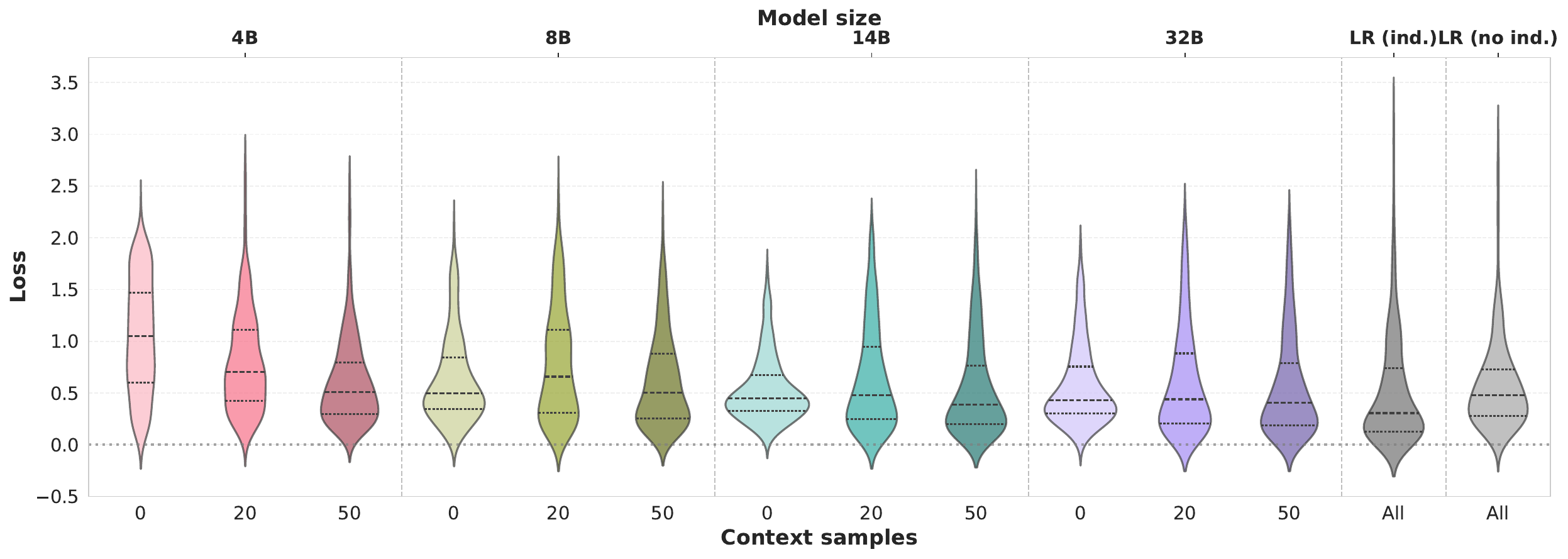}
    \caption{Impact of in-context learning on individual log-loss as a function of model size and number of in-context samples. All models use serialized missingness without steering (Indicator). \textit{As the context includes more samples, the predictive distribution shifts closer to the logistic regression baseline with missingness indicators, demonstrating how the LLM may be learning patterns to modify their probabilistic beliefs.}}
    \label{fig:mimic:icl:logloss}
\end{figure*}

\paragraph{Findings.} Table~\ref{tab:loss} reports the mean log-loss, and Figure~\ref{fig:mimic:serialization:logloss} illustrates the distribution of individual log-loss under the different interventions. Focusing on explicitly serializing missing values in the prompt and its associated representation gain (left panel), we find that explicit missingness representation systematically alters the LLM's predictive beliefs at the individual-sample level, as evidenced by the spread of the log-loss difference. However, the directionality of this effect is highly heterogeneous, with not all patients benefiting from the intervention. On average, all models present a negligible representation gain. This suggests that while the model accounts for missingness, it struggles to leverage it to reduce predictive error without further steering in the zero-shot setting. Interestingly, the largest and smallest models exhibit the largest spread, reflecting greater sensitivity to missingness serialization.

\subsection{Instruction: Zero-shot LLMs Can Be Steered}
\label{sec:zero-shot-inst} 
\paragraph{Intervention.} We evaluate two distinct steering instructions, denoted as \textbf{Steered (Implicit)} and \textbf{Steered (Explicit)}. By incorporating $I$, we provide the model with a formal prior to interpret and leverage feature-missingness patterns during prediction. Details on prompts are provided in App.~\ref{app:prompt}.
\begin{asparaitem}
    \item \textbf{Steered (Implicit):} The instruction prompts the LLM to infer potential missingness informativeness from context.
    \item \textbf{Steered (Explicit):} The instruction provides explicit prior knowledge, describing the relation between missingness and outcome (e.g., intentional omission of a test reflects a patient's stability). 
\end{asparaitem}
\paragraph{Findings.} Focusing on the steering gain in Figure~\ref{fig:mimic:serialization:logloss} and associated log-loss in Table~\ref{tab:loss}, we find that the steering intervention systematically reduces log-loss across various model sizes for the explicit variant, simultaneously improving the average expected loss and shifting the overall error distribution. We observe a general trend: increasing model size leads to lower expected losses in the zero-shot setting, with the 14B model achieving the lowest overall log-loss after steering. This provides evidence that LLMs can incorporate structural constraints via natural-language instructions, effectively guiding the selection of a functional form. We validate these findings with Gemma models in App.~\ref{app:domain_gemma}. However, when the instruction is implicit, LLMs are unable to infer the impact of missingness on the outcome from their prior knowledge, and steering gain is not consistently achieved. We find in App.~\ref{app:ablation_infer} that larger models tend to increase risk under the implicit instruction, steering the predictive distribution towards a region unaligned with the true target distribution.

\begin{table}[!ht]
\centering
\caption{Cohort log loss (mean and 95\% CI) by model size, prompt variant, and context length.}
\label{tab:loss}
\resizebox{\columnwidth}{!}{%
\begin{tabular}{llccc}
\toprule \textbf{Size} & \textbf{Prompt Variant} & \textbf{0-shot} & \textbf{20-shot} & \textbf{50-shot} \\ \midrule 
\multirow{4}{*}{4B}
& Dropped & 1.069 {\small (1.019, 1.118)} & 0.944 {\small (0.887, 1.000)} & 0.836 {\small (0.782, 0.890)} \\
 & Indicator & 1.047 {\small (0.999, 1.095)} & 0.798 {\small (0.757, 0.840)} & 0.602 {\small (0.566, 0.638)} \\
 & Steered (Implicit) & 0.878 {\small (0.836, 0.920)} & 0.515 {\small (0.490, 0.539)} & 0.405 {\small (0.376, 0.433)} \\
 & Steered (Explicit) & 0.836 {\small (0.789, 0.882)} & 0.500 {\small (0.475, 0.525)} & 0.391 {\small (0.361, 0.421)} \\
\midrule
\multirow{4}{*}{8B} & Dropped & 0.618 {\small (0.582, 0.654)} & 0.781 {\small (0.734, 0.828)} & 0.672 {\small (0.628, 0.715)} \\
 & Indicator & 0.633 {\small (0.596, 0.669)} & 0.775 {\small (0.728, 0.822)} & 0.613 {\small (0.574, 0.652)} \\
 & Steered (Implicit) & 0.609 {\small (0.574, 0.644)} & 0.691 {\small (0.650, 0.733)} & 0.557 {\small (0.522, 0.593)} \\
 & Steered (Explicit) & 0.507 {\small (0.473, 0.542)} & 0.624 {\small (0.583, 0.665)} & 0.510 {\small (0.476, 0.544)} \\
\midrule
\multirow{4}{*}{14B} & Dropped & 0.533 {\small (0.504, 0.562)} & 0.663 {\small (0.617, 0.709)} & 0.579 {\small (0.535, 0.624)} \\
 & Indicator & 0.534 {\small (0.506, 0.562)} & 0.640 {\small (0.596, 0.683)} & 0.549 {\small (0.508, 0.591)} \\
 & Steered (Implicit) & 0.568 {\small (0.538, 0.598)} & 0.712 {\small (0.668, 0.756)} & 0.574 {\small (0.535, 0.613)} \\
 & Steered (Explicit) & 0.426 {\small (0.397, 0.456)} & 0.588 {\small (0.545, 0.631)} & 0.515 {\small (0.472, 0.557)} \\
\midrule
\multirow{4}{*}{32B} & Dropped & 0.595 {\small (0.559, 0.631)} & 0.616 {\small (0.570, 0.662)} & 0.548 {\small (0.505, 0.590)} \\
 & Indicator & 0.572 {\small (0.538, 0.606)} & 0.606 {\small (0.561, 0.651)} & 0.563 {\small (0.520, 0.607)} \\
 & Steered (Implicit) & 0.644 {\small (0.608, 0.679)} & 0.612 {\small (0.569, 0.655)} & 0.575 {\small (0.532, 0.618)} \\
 & Steered (Explicit) & 0.489 {\small (0.456, 0.522)} & 0.508 {\small (0.466, 0.550)} & 0.477 {\small (0.435, 0.519)} \\
\bottomrule
\end{tabular}
}
\end{table}

\subsection{In-context learning: Learning missingness patterns from samples}
\label{sec:few-shot-serialization}
\paragraph{Intervention.} We turn to the ICL setting, where the LLM is provided with examples and their associated outcome using patients from the target distribution. Note that we uniformly sample examples from a held-out dataset to maintain the original prevalence of the outcome. We begin by evaluating whether, as we increase the number of context samples from $20$ to $50$, the LLM successfully leverages the observed samples to update its probabilistic beliefs. We then study how explicit steering impacts gain at different context sizes.

\paragraph{Findings.}
Figure~\ref{fig:mimic:icl:logloss} shows the distribution of log-loss as we increase context samples, demonstrating that LLMs across all sizes sharpen probabilistic beliefs as sample size increases. We show in Tables~\ref{tab:loss} and~\ref{tab:ece} in the Appendix that adding context samples within each intervention provides mixed evidence for decreasing average log-loss and ECE, indicating that LLMs do not consistently align probabilistic beliefs with the target distribution.

To understand why, additional analyses reveal failure modes of ICL, where conditioning on context samples induces extreme overconfidence in specific regions of the LLM's predictive distribution. As evidenced by the heavy positive tails in Figure~\ref{fig:mimic:icl:logloss}, the model incurs catastrophic log-loss penalties for a distinct subset of patients. Investigating the relationship between ICL-predicted risk and patient-level $\Delta$ log-loss (Figure~\ref{fig:scatter_delta}) reveals that these penalties are predominantly driven by false positive cases: patients presenting with severe baseline physiology who ultimately survive. Upon observing mortality patterns in the context samples $C_k$, the ICL-conditioned model $q_\psi$ incorrectly assigns high-certainty mortality risk to these matching phenotypes. We interpret this as a failure of the LLM to regularize its predictive function using clinical knowledge or base prevalence. Instead, the model overfits to the context samples, prematurely collapsing its predictive entropy. Consequently, these findings highlight a critical vulnerability of few-shot learning in clinical domains, where unconstrained pattern matching with limited context can lead to estimation errors for specific patient subgroups.

\begin{figure}
    \centering
    \includegraphics[height=140px]{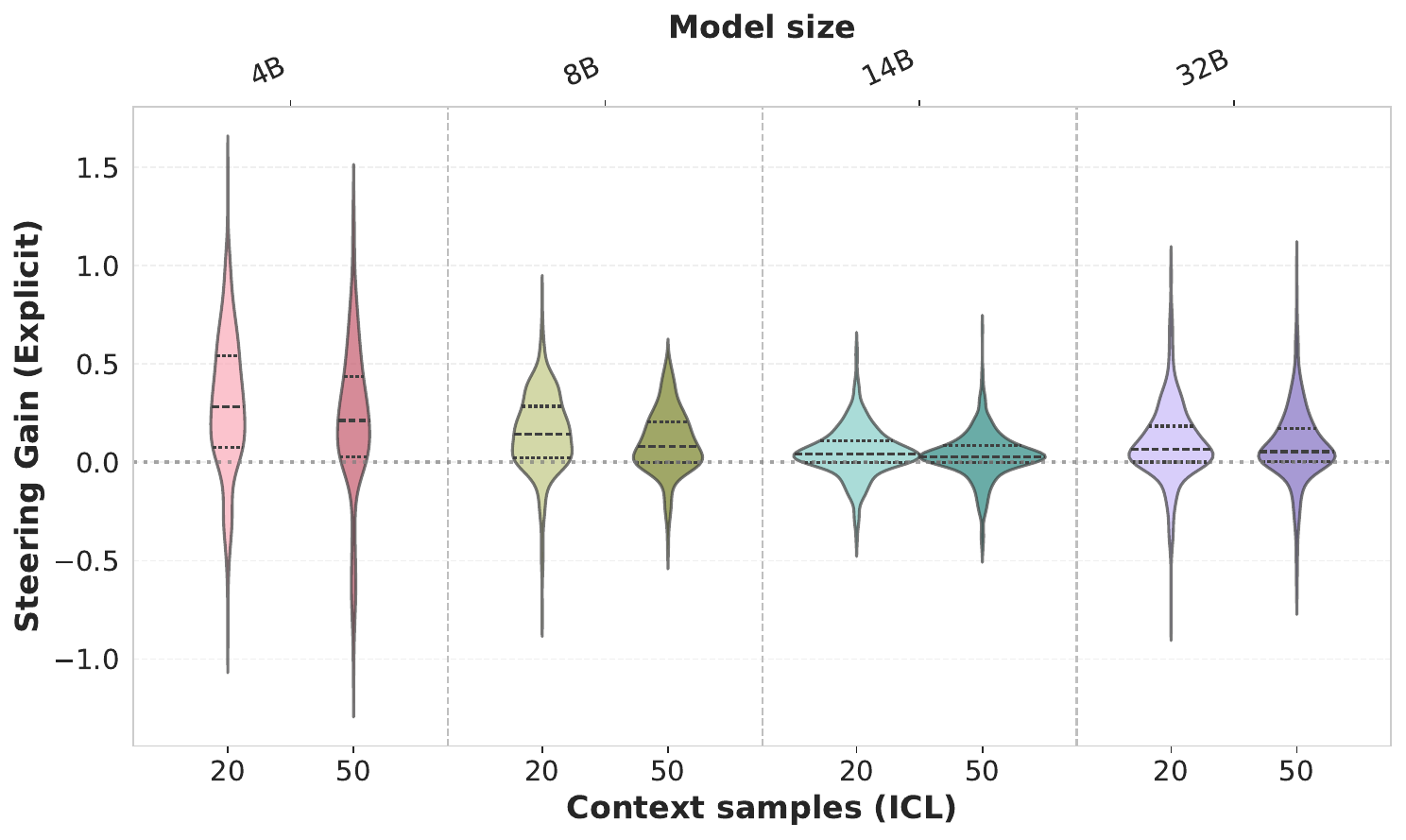}
    \caption{Interaction between ICL and steering given model size and in-context samples. Lighter shade corresponds to $k = 20$ context samples, while darker shade corresponds to $k = 50$ samples. \textit{Additional steering improves performance in ICL settings.}}
    \label{fig:ICL_steering}
\end{figure}

Finally, Figure~\ref{fig:ICL_steering} demonstrates that the performance gains from instruction-based steering are complementary to those of ICL. The best calibration is achieved when the model is simultaneously conditioned on $k = 50$ context samples and the structural steering instruction $I$. This provides evidence that LLMs can successfully leverage few-shot context to capture simple correlative patterns in the target distribution while using explicit natural-language constraints to regularize and calibrate their inferred predictive functions during inference.

\section{Discussion}

Our work demonstrates that general-purpose LLMs are sensitive to missingness, despite their apparent agnosticism toward it. Motivated by the need for reliable uncertainty estimation in the presence of informative missingness, we investigate whether LLMs can refine their probabilistic beliefs by leveraging informative patterns of missingness in zero- and few-shot settings. Our findings reveal that off-the-shelf LLMs generally fail to reliably model these patterns or update their beliefs accordingly, but instruction-based steering helps align the verbalized probabilistic beliefs with the underlying generative process.

We acknowledge that if the objective were solely to obtain calibrated probabilities on a fixed dataset, traditional supervised methods or fine-tuned discriminators would be more suitable. However, our analysis aligns with the prevailing deployment paradigm, in which off-the-shelf LLMs are increasingly used as general-purpose reasoning agents, including in clinical settings. We therefore evaluate whether these models can derive accurate probabilistic beliefs through textual reasoning and prior world knowledge, rather than through task-specific parameter optimization.

While we focus on risk prediction tasks where downstream users are likely to be clinicians or healthcare specialists, this work has implications for patient-facing LLMs. As the general public increasingly relies on LLMs for health advice~\citep{ayre2025use, shahsavar2023user, kullgren2025national}, inconsistent handling of missing information may endanger patients' safety.
For instance, a patient may prompt an LLM with only a subset of the information in a given blood test. In this case, missingness does not reflect a medical process, but the user's medical literacy or behavior. Ensuring that such missingness is accounted for to provide accurate clinical reasoning is therefore critical for users' safety. Our work demonstrates that prompt-based steering offers a path to align probabilistic beliefs with individual target distributions, an opportunity that the traditional machine learning paradigm did not offer, where a model could no longer be applied under such missingness shift~\citep{groenwold2020informative}.

Finally, echoing the criticisms of \cite{nijman2022missing, jeanselme2022imputation} regarding the lack of reporting and inappropriate handling of missing data in machine learning, this work emphasizes the importance of these practices for developing and deploying LLMs. While the paradigm enabled by these models further disconnects training data quality, such as missingness patterns, from downstream performance, our work shows that data quality, specifically missingness, still impacts probabilistic reasoning. 

\paragraph{Limitations.} Our analysis presents evidence of the impact of missingness on LLMs' probabilistic beliefs. The observational nature of our analysis aims to reflect the real-world setting in which these models are used and to evaluate their capacity to leverage contextual information and external knowledge to address non-at-random missingness patterns. However, this approach limits the study of observational missingness, as one cannot enforce realistic missing-at-random or missing-not-at-random data that would be captured in model pretraining, thereby limiting understanding of which types of missingness these models may be robust to. Our reliance on observational outcomes as the ground truth presents two limitations. First, binary outcomes are inherently noisy proxies for evaluating probabilistic beliefs. Second, evaluating reasoning based solely on predictive performance fails to assess faithfulness without expert clinical adjudication of the intermediate steps.

\paragraph{Conclusion.} The proposed analysis offers a crucial insight: the design of LLM applications must pay more attention to the importance of information the user does \emph{not} provide. Our results show that off-the-shelf models are unable to capture informative missingness. However, careful steering can align LLMs' probabilistic beliefs with the underlying data-generating process.

\section{Acknowledgments}
AI-based editing tools were used for language refinement. VJ and SJ would like to acknowledge partial support from NIH 5R01MH137679-02. YK and SJ acknowledge partial support from the RS Fund at Columbia. SJ would like to acknowledge partial support from the Google Research Scholar Award and the SNF Center for Precision Psychiatry \& Mental Health at Columbia. 
Any opinions, findings,  conclusions, or recommendations in this manuscript are those of the authors and do not reflect the views, policies, endorsements, expressed or implied, of any aforementioned funding agencies/institutions.

\bibliography{jmlr-sample}

\newpage
\onecolumn

\title{\titleName\\(Supplementary Material)}
\maketitle

\appendix

\section{Proofs}

\subsection{When does informative missingness occur?}
\label{app:info_mnar}

\begin{theorem}
    Assuming the outcome $Y$ is solely determined by observed features $X$ and potential confounders $U$, the missingness $M$ is informative iff missingness is not at random as defined in~\citet{rubin1976inference}. 
\end{theorem}

\begin{proof}
Let us first prove that, under the Missing At Random (MAR) assumption, then $M \perp Y \mid X_{\text{obs}}$. 

The MAR assumption states that the missingness patterns only rely on observed features:
\begin{equation}
p(M \mid X_{\text{obs}}, X_{\text{mis}}, U) = p(M \mid X_{\text{obs}})
\end{equation}

By Bayes' rule:
\begin{align}
p( X_{\text{mis}}, U \mid X_{\text{obs}}, M) &= \frac{p(M \mid X_{\text{obs}}, X_{\text{mis}}, U) \cdot p( X_{\text{mis}}, U \mid X_{\text{obs}})}{p(M \mid X_{\text{obs}})} \\
&= \frac{p(M \mid X_{\text{obs}}) \cdot p( X_{\text{mis}}, U \mid X_{\text{obs}})}{p(M \mid X_{\text{obs}})} \quad \text{(by MAR)} \\
&= p( X_{\text{mis}}, U \mid X_{\text{obs}})
\end{align}

Thus $M \perp ( X_{\text{mis}}, U) \mid X_{\text{obs}}$. 

Under the assumption that $Y$ is solely determined by $X$ and $U$, we have:
\begin{align}
p(Y \mid X_{\text{obs}}, M) &= \int \int p(Y \mid X_{\text{obs}}, X_{\text{mis}}, U) \cdot p( X_{\text{mis}}, U \mid X_{\text{obs}}, M) \, d X_{\text{mis}} \, dU \\
&= \int \int p(Y \mid X_{\text{obs}}, X_{\text{mis}}, U) \cdot p( X_{\text{mis}}, U \mid X_{\text{obs}}) \, d X_{\text{mis}} \, dU \\
&= p(Y \mid X_{\text{obs}})
\end{align}

Therefore $M \perp Y \mid X_{\text{obs}}$.
\end{proof}

A similar proof ensues under the Missing Completely At Random (MCAR) assumption in which $p(M \mid X_{\text{obs}}, X_{\text{mis}}, U) = p(M)$. Therefore, informative missingness occurs iff MNAR or $M$ is a direct cause of the observed outcome.

\subsection{Proof Theorem \ref{theorem:decomposition}}
\begin{proof}
\begin{align*} &\mathbb{E}_{X,M,C_k}\big[ \mathrm{KL}(p^* \,\|\, q_\psi) \big]
\\
&= \mathbb E_{X,M,C_k}\bigg[-\mathbb{E}_{Y \mid X,M,C_k}\big[\log q_\psi - \log p^* \big] \bigg] \\
&= \mathbb E_{X,M,C_k}\bigg[\mathbb{E}_{Y\mid X,M,C_k}\big[ \log p^* - \log q_\psi + (\log q^* - \log q^*)\big] \bigg] \\
&= \mathbb E_{X,M,C_k}\bigg[\mathbb{E}_{Y\mid X,M,C_k}\big[ (\log p^*  - \log q^*) + (\log q^* - \log q_\psi) \big] \bigg] \\
&= \mathbb E_{X,M,C_k}\bigg[\mathbb{E}_{Y\mid X,M,C_k}\big[\log \frac{p^* }{q^*} + \log \frac{q^*}{q_\psi}\big]\bigg] \\
&= \mathbb E_{X,M}\big[\text{KL}(p^*  \mid \mid q^*)\big]+ \mathbb E_{X,M,C_k}\big[\text{KL}(q^* \mid \mid q_\psi)\big] \end{align*} 
\end{proof}

\subsection{Steering Gain}
\label{app:steering_theory}
We provide a brief discussion on when an informative instruction in the prompt may lead to steering gain. We note that prior work has formalized prompt engineering with learning theory-based arguments \citep{akinwande2023understanding, madras2025prompts}. In contrast, we provide a more general functional intuition. Let $\mathcal{P}$ be the set of all possible finite prompt configurations $\psi = (\phi, I, C_k)$. For this analysis, we consider a prompt family $\mathcal{P}_I = \{(\phi, I', C) \in \mathcal{P} \mid I' = I\}$ which allows the context set to vary but conditions on a specific instruction $I$, leading to a prompt-induced hypothesis class.

\begin{proposition}
    (Complexity Reduction via Steering).  Let $\mathcal{Q}_I$ be the corresponding function class of the prompt family $\mathcal P_I$ with instruction $I$, and $\mathcal{Q}_I \subseteq \mathcal{Q}$. Then it follows from the property of the supremum that $$\widehat{\mathfrak{R}}_k(\mathcal{Q}_I) \leq \widehat{\mathfrak{R}}_k(\mathcal{Q})$$

where $\widehat{\mathfrak{R}}(\mathcal Q)$ is the empirical Rademacher complexity defined as the supremum over the function class associated with $\mathcal Q$, $\widehat{\mathfrak{R}}_k(\mathcal Q) = \mathbb{E}_{\sigma} \left[ \sup_{q \in \mathcal Q} \frac{1}{k} \sum_{i=1}^k \sigma_i q(z_i) \right]$ for a sample set of $k$ examples $\{z_i\}_{i=1}^k$ from a given distribution and $\sigma_i$ are independent random variables drawn from ${\displaystyle \Pr(\sigma _{i}=+1)=\Pr(\sigma _{i}=-1)=1/2} $
\label{prop:complexity}
\end{proposition}

Intuitively, a greater $\widehat{\mathfrak{R}}$ indicates a flexible hypothesis class that can align with randomly generated $\pm1$ labels. If we assume that ICL approximates empirical risk minimization for $C_k$, then using the standard uniform convergence result~(\cite{shalev2014understanding}, Theorem 26.5), we obtain the following upper bound when considering the steered LLM $\mathcal{Q}_I$:

$$\mathbb{E}_{X,M} \big[ \mathbb{E}_{C_k} [ \text{KL}(q_{I^*} \,\|\, q_I) ] \big] \leq 2 \widehat{\mathfrak{R}}_k(\mathcal{Q}_I) + \mathcal{O}\left(\sqrt{\frac{\log(1/\delta)}{k}}\right)$$

where $q_{I^*} := \arg\min_{q \in \mathcal Q} \mathbb{E}_{X,M} [ \text{KL}(p^* \,\|\, q) ]$ is the best possible function within the constrained function class. Note that $\mathcal Q_I$ may not always contain $q^*$ if an instruction is misspecified, such as a factually incorrect instruction with respect to the true data-generating process. However, when it is, the instruction reduces the sample complexity $k$ by Proposition~\ref{prop:complexity}, requiring fewer samples to achieve the same loss.

\newpage
\section{Prompt design}
\label{app:prompt}
\subsection{Instructions}
Our system prompt is dynamically constructed based on the experimental condition (e.g., zero-shot versus few-shot, diffuse instruction versus explicit steering). Variables in brackets, such as \texttt{[COHORT]}, are dynamically populated at inference time with the specific unit (e.g. Coronary Care Unit). The exact text used for our prompt modules is provided below.

\paragraph{Base Prompt and Persona}
All queries begin with the following core persona and task definition, followed by the first step of the reasoning constraint:

\begin{quote}
\textit{You are an expert Clinical Risk Estimation System analyzing a patient record from an emergent [COHORT] admission. Your goal is to estimate the risk of in-[COHORT] mortality based on data collected during the first 48 hours of the [COHORT] stay.}

\textit{Please provide your analysis step-by-step using the following structure:}

\textit{1. CLINICAL ASSESSMENT: Analyze mortality risk based on the observed physiology (demographics, labs, vital signs etc.).}
\end{quote}

\paragraph{Missingness Intervention Modules}
When evaluating the model's ability to process missingness patterns, one of the following two modules is appended to the reasoning structure.

Implicit Instruction:
\begin{quote}
\textit{2. MISSINGNESS MECHANISM: Analyze WHY specific features are missing. Consider whether their absence is potentially informative of the outcome.}
\end{quote}

Explicit Instruction:
\begin{quote}
\textit{2. MISSINGNESS MECHANISM: Recognize that missing values reflect a clinician's decision that the patient is stable. Use the absence of measurements as a protective signal.}
\end{quote}

\paragraph{In-Context Learning}
For few-shot evaluations, we append a pattern recognition instruction. The wording adapts based on whether missingness instructions are active:

With Missingness Instructions:
\begin{quote}
\textit{3. PATTERN RECOGNITION: Look at the few-shot examples provided from the hospital's [COHORT]. Identify any hospital-specific risk patterns and correlations using (a) observed values and (b) whether a feature is measured or not).}
\end{quote}

Without Missingness Instructions:
\begin{quote}
\textit{2. PATTERN RECOGNITION: Look at the few-shot examples provided from the hospital's [COHORT]. Identify any hospital-specific risk patterns and correlations using observed values.}
\end{quote}

\paragraph{Output Constraint}
All prompts conclude with the following strict formatting constraint to ensure reliable extraction of the continuous probability:

\begin{quote}
\textit{After your analysis, you must output the final probability in a strictly valid JSON block at the very end of your response. Use this format:}
\begin{verbatim}
```json
{
  "prediction_prob": 0.0 to 1.0
}'''
\end{verbatim}
\end{quote}

\subsection{Laboratory test serialization}
Following the previous instruction, we include the patient's laboratory test results via serialization. The latter consists of listing all results in a textual format. The example shows the missingness indicator serialization strategy applied to a synthetically generated patient with a subset of measurements (for illustrative purposes). 

\begin{verbatim}
# Electronic Health Record
## Demographics
Patient age: 65.2
Patient gender: M
## Most Recent Measurements
- Heart Rate
  - 88.00
- Mean BP
  - 70.00
- SpO2
  - 96.00
- Creatinine
  - 1.20
- BUN
  - 20.00
- WBC
  - 12.50
- Lactate
  - Not measured
- Troponin I
  - Not measured
\end{verbatim}

\subsection{Sample responses}
We provide excerpts of sample responses in which the LLM verbalized its reasoning about missing features or patterns in the provided samples.

With steering instruction: 
\begin{quote}
    \textbf{\textit{Summary of Risk Factors:}}
        \begin{asparaitem}
            \itshape
            \item The patient has multiple risk factors for in-CCU mortality: advanced age, elevated BUN and creatinine (indicating CKD), hyperglycemia, mild hypotension, and mild respiratory compromise.
            \item The absence of measurements for SaO2, neutrophils, and lymphocytes may reflect clinical stability or lack of active infection, which is a protective factor.
        \end{asparaitem}
\end{quote}

With ICL:
\begin{quote}
    \textbf{From the examples:}
    \begin{asparaitem}
        \itshape
            \item \textbf{High-risk patients typically have}:
            \begin{asparaitem}
              \item  Elevated troponin levels (e.g., Example 3, 5, 16, 43)
              \item  Marked metabolic acidosis (e.g., Example 3, 5, 16)
              \item  Severe renal impairment (e.g., Example 3, 5, 16)
              \item  Elevated INR and PT (e.g., Example 3, 5, 16)
              \item  Severe hypotension (e.g., Example 3, 5)
              \item  Elevated lactate (e.g., Example 3, 5, 16, 43)
              \item  High neutrophil counts (e.g., Example 3, 5, 43)
            \end{asparaitem}

            \item \textbf{Low-risk patients typically have}:
            \begin{asparaitem}
              \item Normal or mildly elevated troponin (e.g., Example 1, 2, 4, 6, 7, 8, 9, 10, 11, 12, etc.)
              \item Normal or only slightly abnormal labs
              \item No severe acidosis or hypotension
            \end{asparaitem}

            \item \textbf{Pattern Matching}:
            \begin{asparaitem}
                \item This patient has elevated troponin, mild metabolic acidosis, mild renal impairment, mildly elevated lactate, and elevated neutrophils.
                \item These findings align with high-risk patients seen in the examples (e.g., Example 3, 5, 16, 43).
                \item However, the absence of severe hypotension, severe acidosis, or markedly elevated INR/PT suggests that the risk is moderate rather than severe.
            \end{asparaitem}
    \end{asparaitem}
\end{quote}

\subsection{Representation and steering gains}
\label{app:test_statistic}
While $\mathcal R(s_{\text{exp}})$ and $\mathcal{S}(I)$ are defined theoretically via KL divergence in Section~\ref{sec:method}, it is empirically computable as the reduction in Cross-Entropy Loss. Since the entropy of the true distribution $H(p^*)$ is constant regardless of the model, the representation and steering gain is equivalent to the improvement: 
$$\mathcal{R}(s_{\text{exp}}) \equiv \mathbb{E}_{Y\mid X,M}\big[-\log q_{s_{\text{imp}}} \big] - \mathbb{E}_{Y\mid X,M}\big[-\log q_{s_{\text{exp}}}\big]$$
$$\mathcal{S}(I) \equiv \mathbb{E}_{Y\mid X,M}\big[-\log q_\emptyset \big] - \mathbb{E}_{Y\mid X,M}\big[-\log q_I\big]$$
This allows us to estimate $\mathcal{R}(s_{\text{exp}})$ and $\mathcal{S}(I)$ without knowing the true distribution $p^*$.

\subsection{Expected calibration error}
Let $Y\in \{0,1\}$ be the true label, and $\hat{p}\in[0,1]$ be the verbalized risk estimate of $Y=1$. We define bins $B_k, k\in{1,...,}K$ that uniformly partitions $[0,1]$. The ECE is computed as follows:
\begin{equation}
\text{ECE} = \sum_{k=1}^{K} \Pr(\hat{p} \in B_k) \, 
\Big| \mathbb{E}[Y \mid \hat{p} \in B_k] - \mathbb{E}[\hat{p} \mid \hat{p} \in B_k] \Big|
\end{equation}
Note that as the bin widths approach 0, the ECE estimates the expected absolute difference between predicted probability and true conditional probability under the distribution of predicted probabilities $\mathbb{E}_{\hat{p}} \Big[ \big| \mathbb{E}[Y \mid \hat{p}] - \hat{p} \big| \Big]$

\newpage
\section{Additional results}

\subsection{Logistic Regression}
\label{app:lr_baseline}
To establish a well-calibrated baseline, we derive predicted probabilities from a Logistic Regression model. We employ 5-fold cross-fitting to generate out-of-fold predictions for the entire dataset. The model is fit under two input conditions: standard mean imputation, and mean imputation augmented with missingness indicators. This comparison serves as a practical proxy to quantify the predictive signal gained by making the missingness mechanism explicitly available to the logistic regression.

Table~\ref{tab:lr_full} demonstrates that using both the full cohort and using 20/50 samples (same context samples as ICL), missingness provides an informative signal.

\begin{table}[!ht]
\centering
\caption{Logistic regression with 20 and 50 training samples (stratified sampling, seeds 1--5) and full dataset (k-fold): log loss and ECE on fixed test set (95\% CI over seeds/fold).}
\label{tab:lr_full}
\begin{tabular}{llcc}
\toprule
 & $n_{\mathrm{train}}$ & Log loss & ECE \\
\midrule
& 20 & 0.602 (0.536, 0.678) & 0.297 (0.257, 0.341) \\
With indicators & 50 & 0.546 (0.476, 0.579) & 0.242 (0.203, 0.270) \\
 & All & 0.518 (0.498, 0.538) & 0.276 (0.262, 0.290) \\
\midrule
 & 20 & 0.627 (0.576, 0.705) & 0.311 (0.275, 0.354) \\
Without indicators & 50 & 0.581 (0.523, 0.614) & 0.262 (0.229, 0.291) \\
 & All & 0.579 (0.563, 0.596) & 0.307 (0.293, 0.321) \\
\bottomrule
\end{tabular}
\end{table}

\begin{figure}[!ht]
    \centering
    \includegraphics[width=0.9\linewidth]{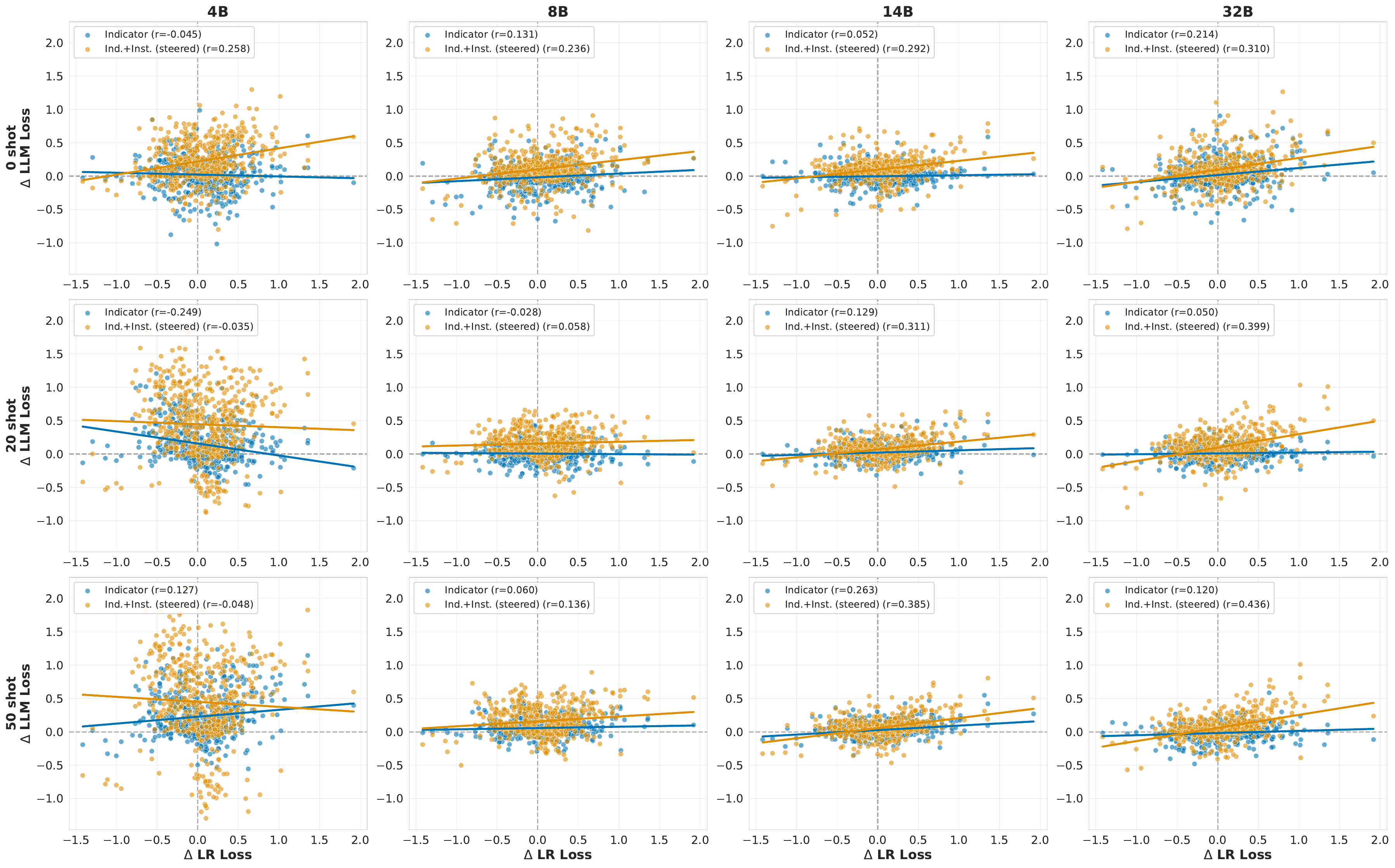}
    \caption{Correlation between the change in loss of the Logistic Regression (LR) and the LLM. \textit{Explicit steering produces a positive correlation: samples that benefit from missingness under logistic regression also benefit under the LLM."}}
    \label{fig:correlation_gain_lr}
\end{figure}

Finally, we compare the change in individual log-loss between the Logistic Regression (LR) and the LLM. Figure~\ref{fig:correlation_gain_lr} evidences a correlation between the change in log-loss, indicating that the steering benefits the same samples as explicit addition of a missingness indicator for the logistic regression. However, the correlations associated with missingness serialization are near zero, demonstrating that the LLMs do not leverage the missingness process under this intervention.

\newpage
\subsection{Tabular results}
\label{app:tab_res}
We compute the ECE for all interventions. Table~\ref{tab:ece} shows consistent improvement as the number of context samples increases and when an explicit steering instruction is included. 

\begin{table}[!ht]
\centering
\caption{Cohort ECE (mean and 95\% CI, bootstrap) by model size, prompt variant, and context length.}
\label{tab:ece}
\begin{tabular}{llccc}
\toprule \textbf{Size} & \textbf{Prompt Variant} & \textbf{0-shot} & \textbf{20-shot} & \textbf{50-shot} \\ \midrule 
\multirow{4}{*}{4B} & Dropped & 0.544 {\small (0.517, 0.575)} & 0.479 {\small (0.449, 0.507)} & 0.427 {\small (0.396, 0.456)} \\
 & Indicator & 0.539 {\small (0.505, 0.566)} & 0.419 {\small (0.386, 0.449)} & 0.319 {\small (0.293, 0.345)} \\
 & Steered (Implicit) & 0.470 {\small (0.441, 0.494)} & 0.264 {\small (0.238, 0.290)} & 0.157 {\small (0.134, 0.185)} \\
 & Steered (Explicit) & 0.446 {\small (0.421, 0.472)} & 0.248 {\small (0.221, 0.275)} & 0.145 {\small (0.123, 0.174)} \\
\midrule
\multirow{4}{*}{8B}  & Dropped & 0.318 {\small (0.287, 0.347)} & 0.406 {\small (0.375, 0.434)} & 0.347 {\small (0.318, 0.374)} \\
 & Indicator & 0.328 {\small (0.297, 0.359)} & 0.406 {\small (0.375, 0.436)} & 0.321 {\small (0.292, 0.348)} \\
 & Steered (Implicit) & 0.316 {\small (0.289, 0.346)} & 0.363 {\small (0.333, 0.390)} & 0.284 {\small (0.255, 0.313)} \\
 & Steered (Explicit) & 0.245 {\small (0.216, 0.272)} & 0.321 {\small (0.293, 0.346)} & 0.255 {\small (0.225, 0.282)} \\
\midrule
\multirow{4}{*}{14B}  & Dropped & 0.260 {\small (0.229, 0.289)} & 0.338 {\small (0.308, 0.370)} & 0.284 {\small (0.257, 0.311)} \\
 & Indicator & 0.260 {\small (0.233, 0.289)} & 0.330 {\small (0.302, 0.357)} & 0.268 {\small (0.234, 0.296)} \\
 & Steered (Implicit) & 0.283 {\small (0.251, 0.311)} & 0.371 {\small (0.340, 0.400)} & 0.287 {\small (0.258, 0.314)} \\
 & Steered (Explicit) & 0.177 {\small (0.149, 0.202)} & 0.291 {\small (0.263, 0.321)} & 0.239 {\small (0.207, 0.269)} \\
\midrule
\multirow{4}{*}{32B}  & Dropped & 0.297 {\small (0.266, 0.326)} & 0.303 {\small (0.276, 0.336)} & 0.258 {\small (0.229, 0.290)} \\
 & Indicator & 0.285 {\small (0.257, 0.316)} & 0.301 {\small (0.270, 0.334)} & 0.272 {\small (0.241, 0.300)} \\
 & Steered (Implicit) & 0.336 {\small (0.307, 0.368)} & 0.307 {\small (0.278, 0.335)} & 0.280 {\small (0.250, 0.310)} \\
 & Steered (Explicit) & 0.225 {\small (0.197, 0.253)} & 0.231 {\small (0.202, 0.262)} & 0.206 {\small (0.177, 0.237)} \\
\bottomrule
\end{tabular}
\end{table}

\newpage
\subsection{Domain-specific Finetuning}
\label{app:domain_gemma}

\begin{figure*}[!ht]
    \centering
    \setlength{\tabcolsep}{1pt} 
    \begin{tabular}{ccc}
        \includegraphics[height=140px]{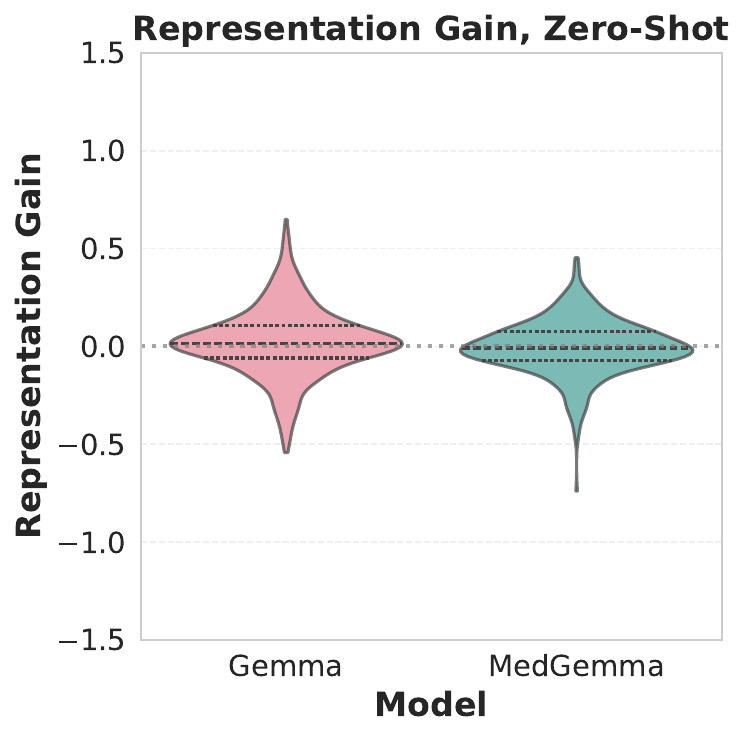} & 
        \includegraphics[height=140px]{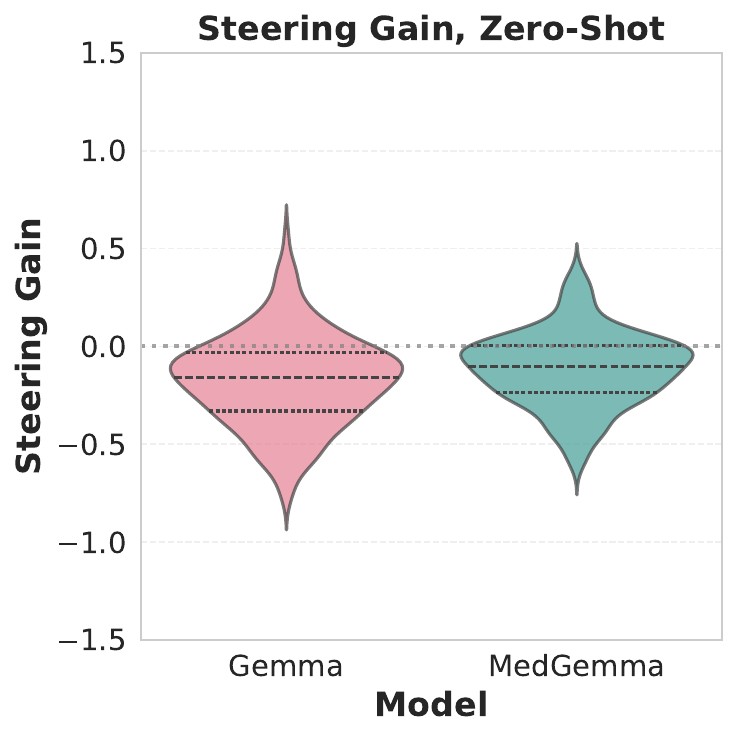} &
        \includegraphics[height=140px]{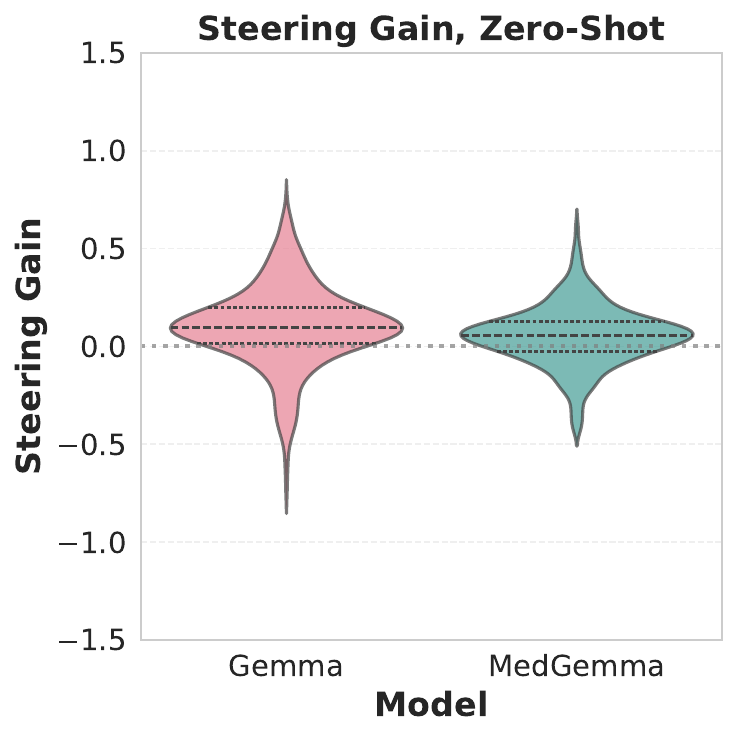} \\
        \small (a) Representation gain & \small (b) Implicit steering gain & \small (c) Explicit steering gain  \\
    \end{tabular}
    \caption{Impact of missingness serialization and instruction steering on individual log-loss difference as a function of model size. \textit{Positive values correspond to improved log-loss under the intervention compared to the base prompt.} }
    \label{fig:logloss_gemma}
\end{figure*}

We evaluate our findings for zero-shot inference on a LLM model family using Gemma 3 (27B) and additionally assess whether clinical-domain-specific fine-tuning can reduce predictive error by improving prior alignment. Note that these two models are the same architecture, with Medgemma \citep{sellergren2025medgemma} further finetuned on medical data. Tables~\ref{tab:gemma_zeroshot_loss} and \ref{tab:gemma_zeroshot_ece} present the log-loss and ECE under the different interventions for these two models. Figure~\ref{fig:logloss_gemma} presents the relative gain.

We find similar patterns to those in the main text's results: explicit steering is required for the LLM to leverage informative missingness, and adding indicators or providing implicit instructions does not consistently improve verbalized beliefs. Interestingly, we find that the standard Gemma model consistently produces calibrated probabilistic beliefs.

\begin{table}[!ht]
\centering
\caption{Zero-shot cohort log loss (mean and 95\% CI) by model and prompt variant.}
\label{tab:gemma_zeroshot_loss}
\begin{tabular}{lcc}
\toprule
\textbf{Prompt Variant} & \textbf{Gemma} & \textbf{MedGemma} \\
\midrule
Dropped & 0.640 {\small (0.607, 0.672)} & 0.643 {\small (0.613, 0.674)} \\
Indicator & 0.617 {\small (0.587, 0.648)} & 0.649 {\small (0.618, 0.680)} \\
Steered (Implicit) & 0.788 {\small (0.752, 0.823)} & 0.763 {\small (0.727, 0.798)} \\
Steered (Explicit) & 0.516 {\small (0.483, 0.549)} & 0.595 {\small (0.564, 0.627)} \\
\bottomrule
\end{tabular}
\end{table}

\begin{table}[!ht]
\centering
\caption{Zero-shot cohort ECE (mean and 95\% CI, bootstrap) by model and prompt variant.}
\label{tab:gemma_zeroshot_ece}
\begin{tabular}{lcc}
\toprule
\textbf{Prompt Variant} & \textbf{Gemma} & \textbf{MedGemma} \\
\midrule
Dropped & 0.332 {\small (0.301, 0.363)} & 0.338 {\small (0.306, 0.366)} \\
Indicator & 0.317 {\small (0.283, 0.344)} & 0.340 {\small (0.308, 0.367)} \\
Steered (Implicit) & 0.417 {\small (0.387, 0.442)} & 0.409 {\small (0.380, 0.438)} \\
Steered (Explicit) & 0.237 {\small (0.210, 0.263)} & 0.310 {\small (0.281, 0.338)} \\
\bottomrule
\end{tabular}

\end{table}

\newpage
\subsection{Failure modes}
\label{app:ablation_infer}
\begin{figure}[!ht]
    \centering
    \includegraphics[width=0.9\linewidth]{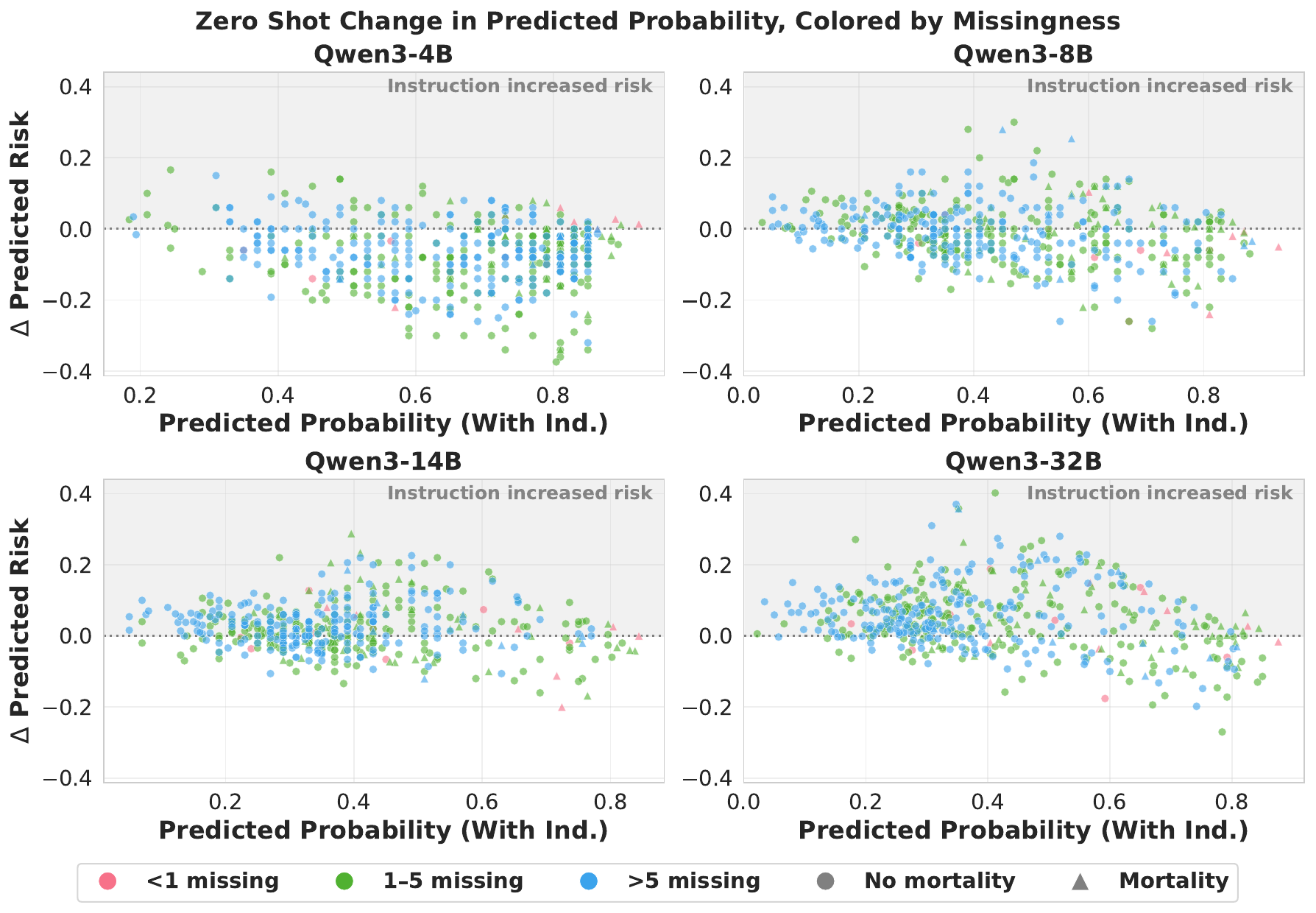}
    \caption{{Change in predicted risk induced by the steering instruction.}\textit{Whereas smaller models exhibit varied shifts, larger models systematically inflate their predicted risk across all missingness subgroups.}}
    \label{fig:scatter_prob_steering}
\end{figure}

In the zero-shot setting, the implicit instruction induces substantial variance in individual-level log-loss change, indicating a significant but highly heterogeneous influence on model predictions. As shown in Figure~\ref{fig:scatter_prob_steering}, increasing model size under steering instruction shifts the direction of predicted risk. Smaller models systematically reduce their predicted risk, whereas larger models tend to inflate it. This divergent behavior persists regardless of the absolute number of missing features. Clinically, the absence of a lab order is typically protective, signaling physiological stability. Consequently, the behavior of larger zero-shot models reveals a misalignment with this true data-generating mechanism: rather than recognizing missingness as a proxy for stability, they become uncalibrated, inflating risk estimates for stable patients while simultaneously heightening confidence for severe cases.

\begin{figure}[!ht]
    \centering
    \includegraphics[width=0.9\linewidth]{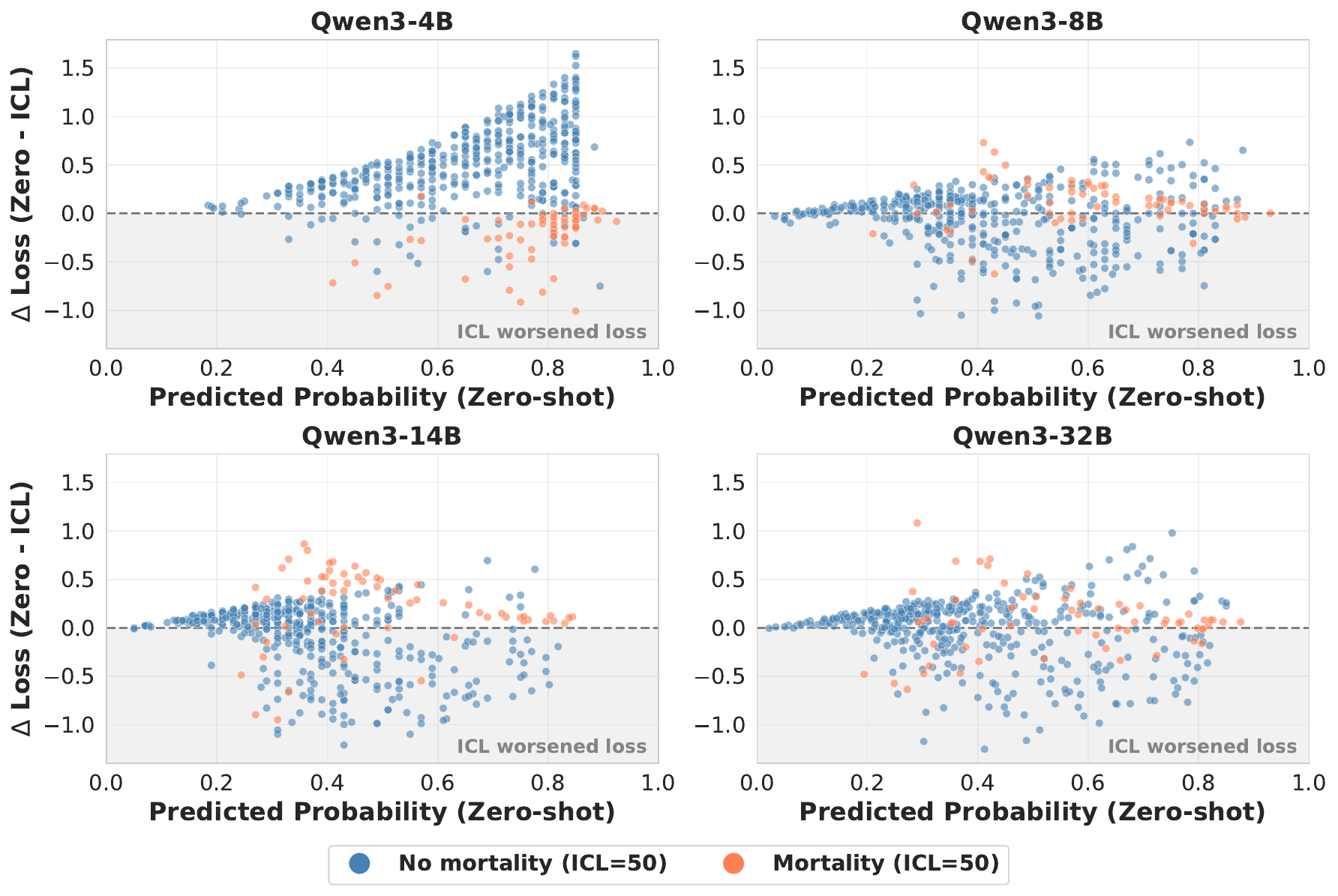}
    \caption{{Change in predicted risk induced by ICL.}}
    \label{fig:scatter_delta}
\end{figure}

\newpage
Similarly, we visualize in Figure~\ref{fig:scatter_delta} the patient-level change in loss when adding 50 context samples, compared to zero-shot predictions (on the x-axis). As model size increases, we observe highly heterogeneous predictive shifts, with significant differences between the positive and negative classes. This variance suggests that the LLM is leveraging the context to perform conditional inference rather than a simple global adjustment. ICL with larger models also demonstrates a failure mode for overconfidence in "false" negative cases.

\end{document}